\documentclass[10pt,twocolumn,letterpaper]{article}

\usepackage{cvpr}              

\usepackage{algorithm}
\usepackage{algpseudocode}
\usepackage{multirow}

\renewcommand{\paragraph}[1]{\vspace{0.2em}\noindent\textbf{#1.}}

\definecolor{cvprblue}{rgb}{0.21,0.49,0.74}
\usepackage[pagebackref,breaklinks,colorlinks,allcolors=cvprblue]{hyperref}

\def\confName{CVPR}
\def\confYear{2027}

\title{Controllable CAD Generation and Interactive Editing of Fluid Rotors}
\author{Shen Chen\textsuperscript{1}, Lei Li\textsuperscript{2}, Jiajun Sun\textsuperscript{1}, Zitian Wang\textsuperscript{1}, Peng Wu\textsuperscript{1}\thanks{Corresponding author.}\\
\textsuperscript{1}Zhejiang University, College of Energy Engineering\\
\textsuperscript{2}Beijing Institute of Technology, School of Artificial Intelligence\\
{\tt\small \{\href{mailto:shenchen3@zju.edu.cn}{shenchen3}, \href{mailto:wu_peng@zju.edu.cn}{wu\_peng}\}@zju.edu.cn, \{\href{mailto:lilei@bit.edu.cn}{lilei}\}@bit.edu.cn}
}
\newif\ifVaneCraftSupplement
\VaneCraftSupplementfalse
\newcommand{\VaneCraftTeaser}{%
    \begingroup
    \centering
    \includegraphics[width=\textwidth,height=34mm,keepaspectratio]{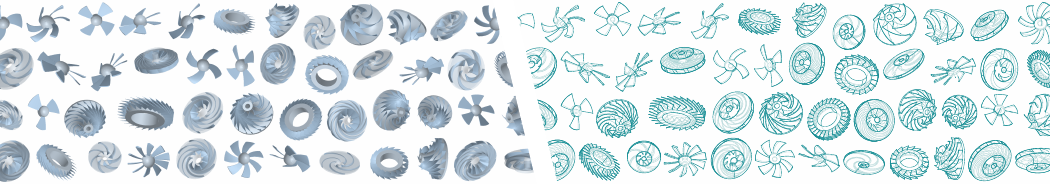}\par
    \captionof{figure}{Fluid Rotors generated by VaneCraft. Shaded renderings and corresponding edge views illustrate geometric diversity across blade configurations.}
    \label{fig:teaser}
    \endgroup
    \vskip 12pt
}
\makeatletter
\g@addto@macro\@maketitle{\VaneCraftTeaser}
\makeatother

\begin{document}
\maketitle
\begin{abstract}
Computer-aided design (CAD) provides the geometric foundation for rotors in fluid rotating machinery.
However, creating models for different rotors requires different modeling processes, making it difficult to control geometry through design intent and reuse existing models.
We introduce VaneCraft, an agent-driven framework for rotor generation, inverse recovery, and editing through a unified analytical representation.
The agent interprets design requests and invokes geometric modeling operations.
The representation defines geometric controls through meridional geometry and independent boundary angle laws.
An anisotropic partial differential equation (PDE) maps these controls to the blade surfaces.
Geometric parameter correction (GPC) uses a joint geometric encoder and a parameter decoder to refine controls initialized from rotor category and dimensions.
Inverse recovery maps existing CAD to the same geometric controls for reconstruction and continued editing.
Experiments demonstrate that VaneCraft supports anisotropy control and parameter editing and outperforms representative baselines in rotor CAD generation tasks.
Project page is available on \href{https://github.com/shenchen333/vanecraft}{GitHub}.
\end{abstract}
    
\section{Introduction}
\label{sec:intro}

\begin{figure*}[t]
    \centering
    \includegraphics[width=0.95\linewidth]{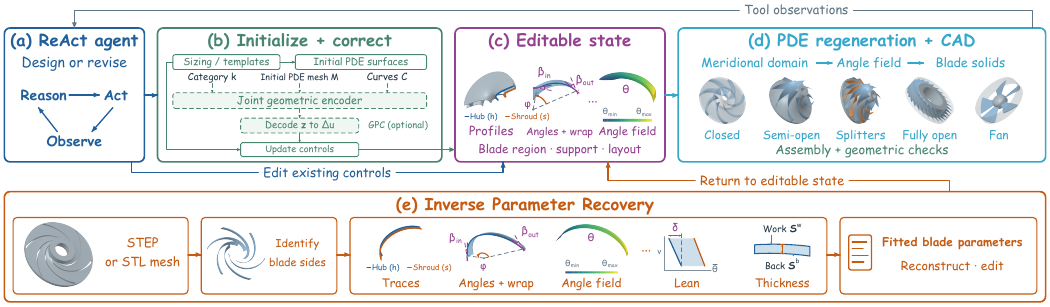}
    \caption{Overview of VaneCraft. (a) An agent selects design or editing operations. (b) Sizing or templates initialize controls for an initial PDE construction. Optional GPC corrects the controls. (c) The editable state drives (d) PDE surface regeneration, CAD assembly, and geometric checks. (e) Parameters recovered.}
    \label{fig:framework}
    \label{fig:shared-geometry}
\end{figure*}

The geometry of rotors in fluid rotating machinery is controlled through meridional profiles, blade angles, thickness, and blade count~\cite{gulich2020,han2018energies,bonaiuti2006analysis,fan2020parametric}.
As an alternative to this design-parameter route, generative methods produce 3D models as point clouds~\cite{luo2021diffusion}, meshes~\cite{siddiqui2024meshgpt}, B-reps~\cite{xu2024brepgen}, and parametric operation sequences~\cite{wu2021deepcad}, but these outputs describe geometry without exposing the blade design parameters above.
Rotor configurations differ in structure and are typically modeled using different parameter definitions and construction rules.
A general blade parametrization supports multiple blade types by fitting scattered surface points~\cite{agromayor2021unified}.
Scattered point coordinates do not explicitly reveal how the spatial arrangement relates to angles, thickness, or meridional profiles.
This makes it difficult to translate design intent into coordinated point movements.

Large language models (LLMs) provide an interface for interpreting design requirements and coordinating CAD tools~\cite{zhang2026llm4cad,zhang2025flexcad,fan2026caddesigner}.
Existing generative CAD research has studied construction sequences, structured shape representations, and text-conditioned modeling~\cite{wu2021deepcad,ganin2021cadlang,xu2022skexgen,xu2023hnc,khan2024text2cad}.
The underlying representations, such as point clouds, meshes, NURBS, B-reps, and parametric operation sequences, differ in how they encode geometry, topology, and construction history, yet all are defined over geometric elements or modeling operations rather than design parameters.
Consequently, LLMs have a hard time understanding the spatial relationships between discrete elements. The connection between their outputs and things like angle, thickness, and coupling constraints is implicit, which makes it tough to control directly through design intent.

We introduce VaneCraft, an agent-driven framework in which rotor generation, inverse recovery, and continued editing operate on a unified analytical representation with explicit geometric controls.
Each blade side is generated from meridional geometry and an angle field governed by an anisotropic Laplace partial differential equation (PDE).
Independent angle laws prescribe the boundary values, while anisotropy adjusts interior angles under fixed boundary values.
Optional geometric parameter correction (GPC) aligns the initial blade mesh and meridional curves. A joint encoder combines these features with the category label, and a decoder predicts control updates.
The same PDE then regenerates the surfaces from the corrected controls.
Assembly operators then realize closed, semi-open, and open rotors with associated casings and fluid domains, as shown in Fig.~\ref{fig:teaser}.

The agent accesses these geometric operators through controllable parameter states.
It selects generation, recovery, and editing operations through a ReAct loop guided by requirements, design context, and tool observations.
Saved states provide anchors for later modifications, while reverse restoration offers an entry point for working from existing geometry. 
As shown in Fig.~\ref{fig:framework}, the construction and inspection process feeds geometric observations back to the agent.

Our contributions are summarized below.
\begin{itemize}
    \item A design framework for fluid rotors in fluid machinery, with a controllable geometric representation that defines design variables and connects generation, recovery, and continued editing.
    \item GPC, a joint encoder--decoder that predicts updates to geometric controls, and an inverse recovery module that maps STEP and STL rotors to the same controls.
    \item A PDE-based representation that combines meridional geometry with angle laws. Anisotropy controls the interior rotor shape while preserving the prescribed boundary values.
\end{itemize}

\vspace{-1pt}

\section{Related Work}

\subsection{Geometry and Parameterized Design}
Parameterized design links geometric controls to surface construction and shape optimization.
PDE methods generate free-form and blade surfaces from prescribed boundaries~\cite{Bloor90,Zhang09TCSAM}.
T-splines~\cite{Sederberg03} introduce T-junctions into the control grid and support local surface refinement.
A unified blade parametrization combines NURBS with design variables for axial, radial, and mixed-flow blades~\cite{agromayor2021unified}.
This method also recovers these variables from scattered surface points through a complex-step method.
Geometric controls support propeller and impeller design~\cite{Dekanski96,Zhang12} and shape optimization~\cite{Li06} under Navier--Stokes flow constraints.
CFD-based parametric studies examine the effects of outlet blade angle on head and efficiency~\cite{Bacharoudis08}.
Surrogate models~\cite{Wang16}, reduced-order models~\cite{Gedda25}, and evolutionary algorithms~\cite{Safikhani11} support the search for improved designs.
VaneCraft supports agent-driven generation, validation, and inverse recovery through a shared parameter state.

\subsection{Generative CAD and Editable Representations}
Structured representations support CAD generation and editing.
Construction sequences encode explicit modeling operations~\cite{wu2021deepcad,ganin2021cadlang,xu2022skexgen,xu2023hnc}.
BrepGen~\cite{xu2024brepgen} generates B-reps through diffusion over hierarchical face, edge, and vertex features.
Text2CAD~\cite{khan2024text2cad} and CAD-GPT~\cite{wang2025cadgpt} generate construction sequences from text.
HoLa~\cite{liu2025hola} supports text-conditioned B-rep generation through holistic latents.
Reusable abstractions simplify shape programs~\cite{ShapeMOD21,ShapeCoder23}, and parameter controls enable interactive editing~\cite{Kodnongbua23,SPAGHETTI22}.
Editability therefore depends on the controls exposed by each representation.

\subsection{Agentic Design and Verification}
Agentic design links language instructions to geometric operations and execution feedback.
CADDesigner~\cite{fan2026caddesigner} combines reasoning, execution, and feedback.
ReAct interleaves reasoning with actions and tool observations~\cite{yao2023react}.
TraceCAD~\cite{fan2026tracecad} uses persistent execution records to localize repairs.
Embodied CAD~\cite{liu2026embodiedcad} guides assembly modeling with solver feedback.
SpatialVLM~\cite{chen2024spatialvlm} and SpatialRGPT~\cite{cheng2024spatialrgpt} infer spatial relations.
ParSEL~\cite{ganeshan2024parsel} propagates language-driven edits through geometric relations to preserve part relationships.
VaneCraft uses geometric feedback to revise design parameters.

\section{Method}
\label{method}

VaneCraft represents a design by the state $\mathcal D=(k,\mathbf p,\boldsymbol\xi)$, where $k$ is the rotor category and $\mathbf p,\boldsymbol\xi$ hold the rotor and optional component parameters.
Closed and semi-open rotors are radial, and fully open rotors are axial.
This state is initialized from design instructions, or recovered from existing CAD.
The PDE representation maps geometric controls to blade surfaces, optional GPC refines initialized controls, and the agent invokes generation, recovery, and editing operations on the same state.

\subsection{Unified Controllable Blade Representation}
\label{pde_method}
\label{sec:representation}
\begin{figure}[t]
    \centering
    \includegraphics[width=0.95\linewidth]{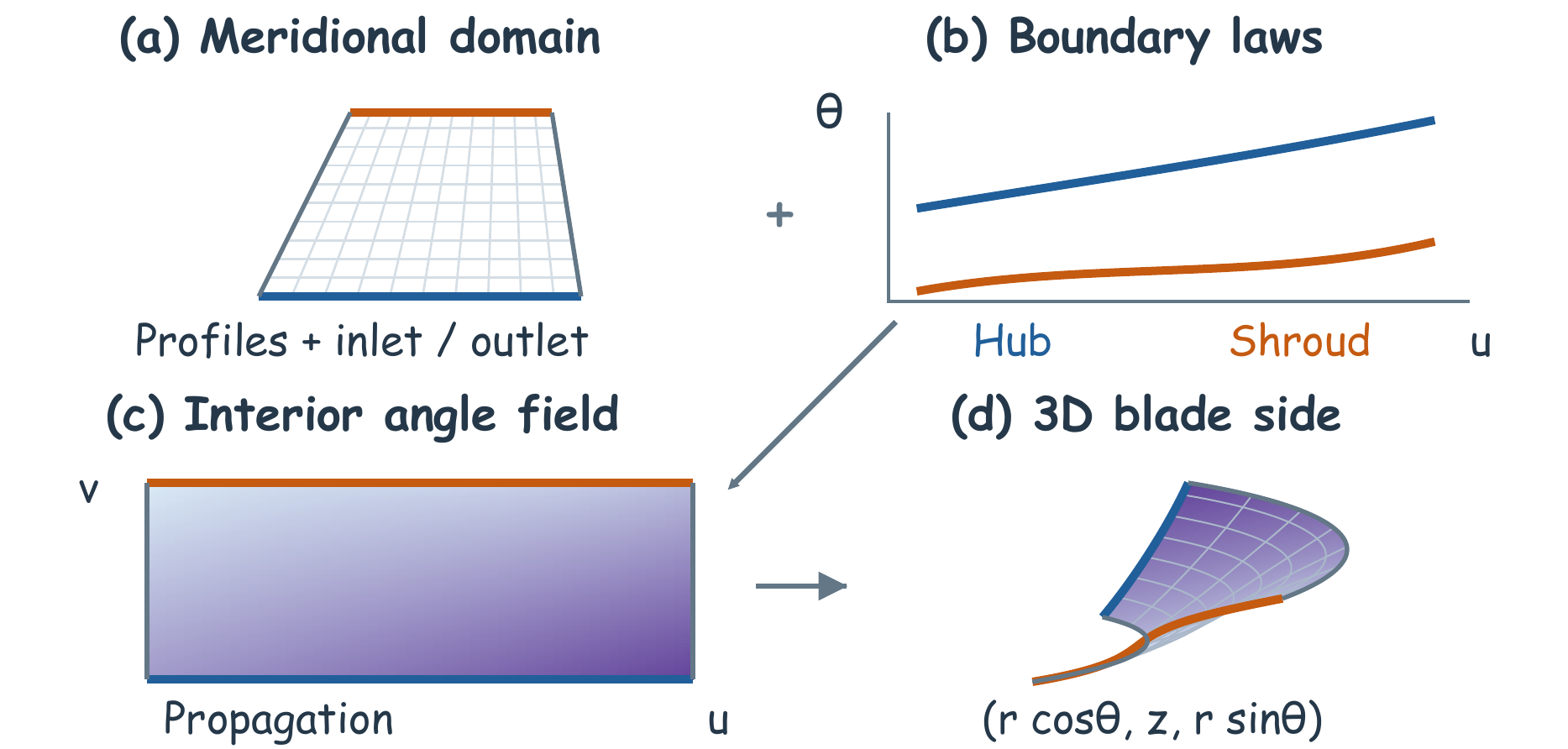}
    \caption{Unified controllable blade representation.  Boundary laws are propagated over the meridional domain to give the 3D blade side.}
    \label{fig:controllable-representation}
\end{figure}

A blade side is described by a meridional map $(u,v)\mapsto(z,r)$, which supplies axial and radial coordinates, and an angle field $\theta(u,v)$, which specifies their circumferential placement, as shown in Fig.~\ref{fig:controllable-representation}.
Both are defined on $\Omega=[0,1]^2$, where $u$ runs along the blade chord from inlet to outlet and $v$ from hub to shroud.
The work side $w$ and the back side $b$ share the meridional map and have separate angle fields.
Sizing formulas in the supplementary material determine the initial dimensions from the design requirements, and predefined control-point ratios initialize the meridional profiles.

\paragraph{Meridional Domain}
The hub and shroud profiles $C_h,C_s$ use B\'ezier curves,
\begin{equation}
    \mathbf C(t)=\sum_{i=0}^{n}\binom{n}{i}
    (1-t)^{n-i}t^i\mathbf P_i,
    \qquad t\in[0,1],
    \label{eq:bezier}
\end{equation}
where $\mathbf P_i=(z_i,r_i)$ are meridional control points, $n$ is the curve degree, and $t$ parameterizes the profile.
Moving the control points adjusts the meridional envelope.
Independent split parameters $t_h,t_s$ select $C_h(t_h)$ and $C_s(t_s)$.
Resampling the downstream intervals $t\in[t_h,1]$ and $t\in[t_s,1]$ by arc length yields the hub segment $\mathbf{h}(u)$ and shroud segment $\mathbf{s}(u)$.
Leading edge $\mathbf{i}(v)$ connects the split points through a radially offset quadratic B\'ezier curve.
Trailing edge $\mathbf{o}(v)$ connects $C_h(1),C_s(1)$. 
These four curves bound the blade domain, and a Coons map defines the meridional coordinates on $\Omega$,
\begin{equation}
\begin{split}
    (z(u,v),r(u,v))&={}(1-v)\mathbf{h}(u)+v\mathbf{s}(u)\\
    &+(1-u)\mathbf{i}(v)+u\mathbf{o}(v)-\mathbf{co}(u,v),
\end{split}
\label{eq:coons}
\end{equation}
where $\mathbf{co}$ is the bilinear interpolation of the four corners.
The shroud profile is a physical boundary for closed rotors and a virtual envelope for unshrouded configurations.

\paragraph{Independent Boundary Laws}
An angle law prescribes the field for each side $q\in\{w,b\}$ and boundary $e\in\{h,s\}$,
\begin{align}
    f_e^q(u)&=\phi_e^q u+\lambda^q g_e^q(u), \label{eq:hub-law}\\
    g_e^q(u)&=a_e^q u-(2a_e^q+b_e^q)u^2
                    +(a_e^q+b_e^q)u^3, \label{eq:correction}
\end{align}
where $a_e^q=\cot\beta_{\mathrm{in},e}^q-\phi_e^q$, $b_e^q=\cot\beta_{\mathrm{out},e}^q-\phi_e^q$, $\phi_e^q$ specifies total wrap, $\beta_{\mathrm{in},e}^q$ and $\beta_{\mathrm{out},e}^q$ are equivalent inlet and outlet angle controls, and $\lambda^q$ sets their influence on the endpoint slopes.
Because $g_e^q(0)=g_e^q(1)=0$, changing $\lambda^q$ preserves total wrap.
The strength $\lambda^q$ interpolates endpoint slopes between the wrap and cotangent controls.

The hub and shroud boundary values are
\begin{equation}
    \theta_h^q(u)=f_h^q(u),\quad
    \theta_s^q(u)=f_s^q(u)+\delta^q .
    \label{eq:lean}
\end{equation}
The shroud inlet phase $\delta^q$ sets the spanwise offset at the inlet, and the difference between shroud and hub wrap adds to it at the outlet.
The inlet and outlet values interpolate linearly between the corresponding hub and shroud endpoints.

\paragraph{Interior Propagation}
The interior angle field solves the anisotropic partial differential equation
\begin{equation}
    \frac{\partial^2\theta^q}{\partial u^2}
    +(\kappa^q)^2\frac{\partial^2\theta^q}{\partial v^2}=0,
    \label{eq:pde}
\end{equation}
with the boundary values from Eqs.~\eqref{eq:lean}.
The coefficient $\kappa^q$ changes the interior field while retaining the prescribed boundary values.
As $\kappa^q\to\infty$, the interior approaches spanwise linear interpolation between the hub and shroud laws, and as $\kappa^q\to0$, chordwise linear interpolation between the inlet and outlet values.
Jacobi relaxation follows the update rule in Algorithm~\ref{alg:blade-generation} on an equally spaced grid, with fixed boundaries and a specified iteration count.
For positive anisotropy, its nonnegative stencil preserves the boundary-value range and converges to the unique finite-grid solution.

\begin{figure}[t!]
    \centering
    \includegraphics[width=0.95\linewidth]{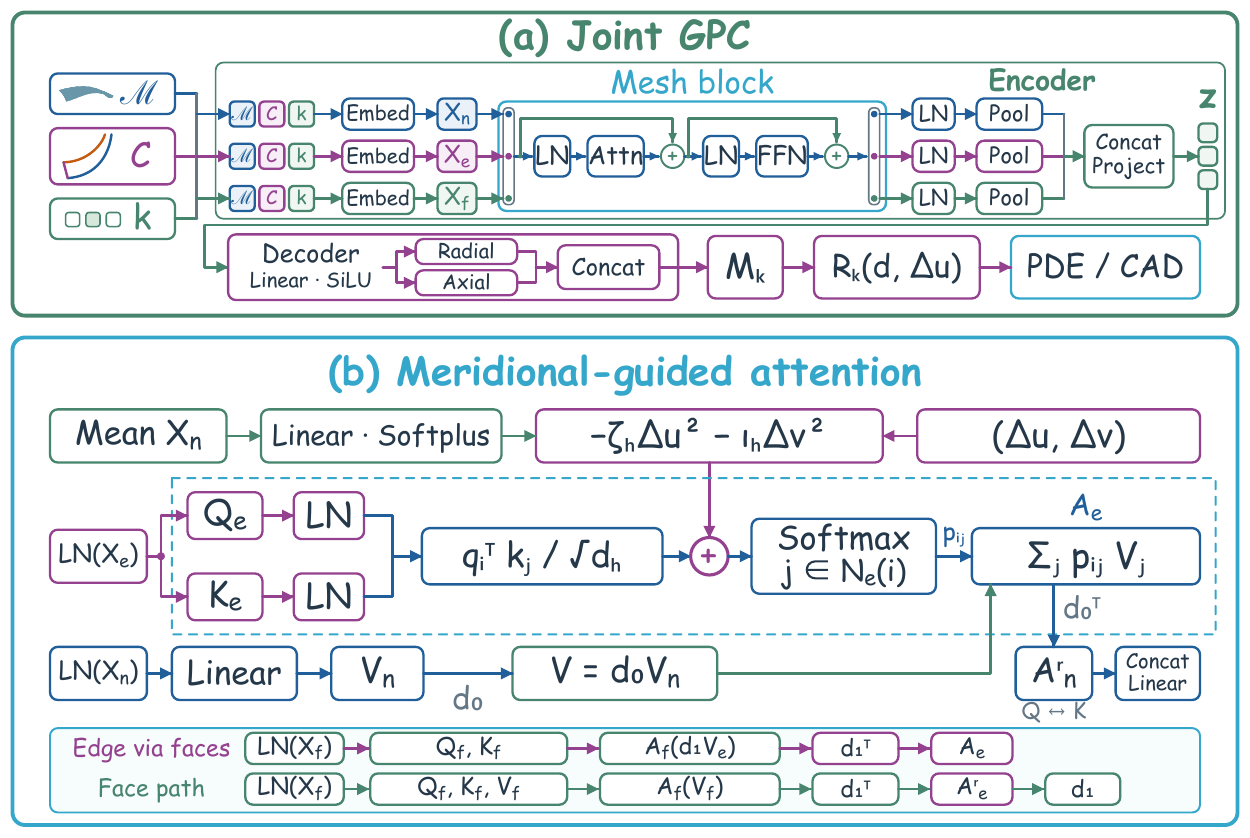}
    \caption{Joint GPC. (a) Embeddings are pooled separately, concatenated, and projected to $\mathbf z$. (b) Meridional-guided attention.}
    \label{fig:correction-editing}
    \label{fig:gpc-overview}
\end{figure}

The solved angle field gives every meridional point $(z,r)$ a circumferential angle and thus a position in space.
In a cylindrical CAD convention with the shaft along $Y$, the side surface $\mathbf{S}^q$ is
\begin{equation}
    \mathbf{S}^q(u,v)=
    \begin{bmatrix}
        r(u,v)\cos\widetilde{\theta}^q(u,v)\\
        z(u,v)\\
        r(u,v)\sin\widetilde{\theta}^q(u,v)
    \end{bmatrix},
    \quad
    \widetilde{\theta}^q=\theta^q-\Delta^q .
    \label{eq:map3d}
\end{equation}
The work and back surfaces $\mathbf{S}^w$ and $\mathbf{S}^b$ have independent angle-law parameters and PDE solves, and their offsets $\Delta^q$ and side laws control the circumferential separation that gives blade thickness.

\subsection{Geometric Parameter Correction}
\label{sec:gpc}
Sizing formulas or templates initialize the controls from category $k$ and dimensions $\mathbf d$ alone.
Optional GPC predicts offsets to the meridional controls from the initial geometry.

\paragraph{Model Architecture}
GPC pairs a joint geometric encoder with a parameter decoder, as illustrated in Fig.~\ref{fig:gpc-overview}(a). 
The encoder takes $k$, meridional curves $C$, and a single-blade mesh $\mathcal M$ triangulated from the initial PDE surfaces, with coordinates normalized by the reference diameter $D$ in $\mathbf d$.
At each element's chordwise and spanwise coordinates $(u,v)$, it concatenates mesh features, one-hot $k$, and hub, shroud, and inlet samples $H(u)$, $S(u)$, and $I(v)$. 
Type-specific linear layers form node, edge, and face embeddings $X_n$, $X_e$, and $X_f$. 
The mesh block combines meridional-guided attention, a feed-forward network (FFN), layer normalization (LN), and residual connections.

The attention follows Hodge--Laplace paths~\cite{nousias2025hodgeformer} along the signed incidence maps $d_0$ from nodes to edges and $d_1$ from edges to faces.
For projected values $V_n$, the node update is $A_n^{\mathrm r}\bigl(d_0^\top A_e(d_0V_n)\bigr)$, where $A_e$ uses $X_e$ queries and keys and $A_n^{\mathrm r}$ swapped queries and keys. The edge and face updates aggregate $d_1V_e$ and $V_f$ with $X_f$ queries and keys.
Before each softmax, weights from mean-pooled $X_n$ penalize squared $(u,v)$ separations, which biases attention toward chordwise and spanwise neighbors, as shown in Fig.~\ref{fig:gpc-overview}(b).
Finally, we concatenate per-type weighted means and maxima and project them to $\mathbf z$.

The decoder $f_{\mathbf W}$ feeds a projection of $\mathbf z$ to radial and axial heads and concatenates their outputs.
The output stage applies the category selector $M_k$ and the constrained map $R_k$ as $    \Delta\mathbf u=M_k[f_{\mathbf W}(\mathbf z)], \widehat{\mathbf p}_{\mathrm{geom}}=R_k(\mathbf d,\Delta\mathbf u),$
where $\Delta\mathbf u$ is the offset vector selected for category $k$ and $\widehat{\mathbf p}_{\mathrm{geom}}$ contains the corrected controls for meridional profiles, inlet shape, and radial blade starts.

\paragraph{Loss}
Training averages a profile loss $d_\partial$, an auxiliary loss $\mathcal L_k$, and a surface loss $d_{\mathrm s}$ across the all categories. The scales $a_k,s_k,b_k$ are their mean training values at zero correction in category $k$, and $\eta,\gamma$ weight the auxiliary and surface terms,
\begin{equation}
    \mathcal L_{\mathrm{train}}=\frac{1}{3}\sum_{k=1}^{3}
    \left(\frac{d_\partial^{(k)}}{a_k}+\eta\frac{\mathcal L_k}{s_k}
    +\gamma\frac{d_{\mathrm{s}}^{(k)}}{b_k}\right),
    \label{eq:correction-training}
\end{equation}

The profile loss fits the hub and shroud profiles. With predicted and reference samples $P_\Gamma,T_\Gamma$ scaled by $D$,
\begin{equation}
\begin{aligned}
d_\partial &= \tfrac14\sum_{\Gamma\in\{h,s\}}
    [\delta(P_\Gamma,T_\Gamma)+\delta(T_\Gamma,P_\Gamma)],\\
\delta(X,Y) &= \tfrac{1}{|X|}\sum_{x\in X}
    \min_{y\in\operatorname{poly}(Y)}\|x-y\|_2 ,
\end{aligned}
    \label{eq:gpc-profile-loss}
\end{equation}
where $\operatorname{poly}(Y)$ joins consecutive samples in $Y$.

The auxiliary loss $\mathcal L_k=\sum_j w_j\,d_j(\widehat G_j,G_j^\star)+\mu_k\mathcal L_{\mathrm{geom}}^{(k)}$ compares normalized predictions $\widehat G_j$ and references $G_j^\star$.
Axial profiles, radial inlet shape, and blade starts use $d_j=\operatorname{MSE}(X,Y)$, the mean squared error between corresponding coordinates. Radial profiles and axial inlets use $d_j=[\delta_2(X,Y)+\delta_2(Y,X)]/2$.
The penalty $\mathcal L_{\mathrm{geom}}^{(k)}=\operatorname{mean}\!\bigl(\max\{0,(r_h-r_s)/D^\star\}^2\bigr)+\operatorname{mean}\!\bigl(\max\{0,2[r_s(0)-r_s(1)]/D^\star\}^2\bigr)$ penalizes $r_h>r_s$ and inlet diameters exceeding outlet diameters.
$r_h,r_s$ denote hub/shroud radii, $0/1$ inlet/outlet, and $D^\star$ the reference outlet diameter. 

The surface loss compares normalized samples $\widehat S,S^\star$ from the reconstructed and aligned reference blades.
\begin{equation}
\begin{aligned}
d_{\mathrm{s}} &= \delta_2(\widehat S,S^\star)+\delta_2(S^\star,\widehat S),\\
\delta_2(X,Y) &= \tfrac{1}{|X|}\sum_{x\in X}
                  \min_{y\in Y}\|x-y\|_2^2 .
\end{aligned}
    \label{eq:gpc-surface-loss}
\end{equation}
The corrected controls drive PDE surface regeneration for CAD assembly in Sec.~\ref{sec:realization}.

\subsection{Structure-Specific CAD Realization}
\label{sec:realization}
Algorithm~\ref{alg:blade-generation} summarizes blade generation from geometric controls to a CAD solid.
B-spline sides and edge closures partition the construction envelope to form a capped blade solid.
Trailing-edge closure follows the cylindrical or noncylindrical layout.

\begin{algorithm}[t]
    \caption{Unified Controllable blade generation}
    \label{alg:blade-generation}
    \small
    \algrenewcommand{\algorithmicrequire}{$\mathcal I$}
    \algrenewcommand{\algorithmicensure}{$\mathcal O$}
    \begin{algorithmic}[1]
        \Require $\mathbf h,\mathbf s,\mathbf i,\mathbf o\in(\mathbb R^2)^{[0,1]},$
        \Statex $\quad\mathcal P=\{\beta_{\mathrm{in},e}^q,\beta_{\mathrm{out},e}^q,\phi_e^q,\delta^q,\Delta^q,\lambda^q,\kappa^q\}_{q\in\{w,b\},\,e=h,s},$
        \Statex $\quad N\geq3,\ T\geq0,\ \kappa^q>0,\ E\subset\mathbb R^3,\ \ell,\ \varepsilon>0.$
        \Ensure $B\in\mathsf{Solids}(E)\cup\{\bot\}.$
        \State $(z,r)\gets\mathsf{Coons}(\mathsf{Resample}_N(\mathbf h,\mathbf s,\mathbf i,\mathbf o))$
        \State $\Omega_N\gets\{(u_i,v_j)=(i,j)/(N-1)\}_{i,j=0}^{N-1}$
        \For{$q\in\{w,b\}$}
            \State $(a_e^q,b_e^q)\gets(\cot\beta_{\mathrm{in},e}^q-\phi_e^q,\ \cot\beta_{\mathrm{out},e}^q-\phi_e^q)$
            \State $f_e^q(u)\gets\phi_e^q u+\lambda^q u(1-u)[a_e^q(1-u)-b_e^q u]$
            \State $\theta_{ij}^{q,0}\gets(1-v_j)f_h^q(u_i)+v_j[f_s^q(u_i)+\delta^q]$
            \For{$t=0,\ldots,T-1$}
                \State $\displaystyle\theta_{ij}^{q,t+1}\gets\frac{\sum_{\sigma=\pm1}\!\left[\theta_{i+\sigma,j}^{q,t}+(\kappa^q)^2\theta_{i,j+\sigma}^{q,t}\right]}{2[1+(\kappa^q)^2]}$
                \Statex $\quad i,j=1,\ldots,N-2$
                \State $\theta^{q,t+1}|_{\partial\Omega_N}\gets\theta^{q,0}|_{\partial\Omega_N}$
            \EndFor
            \State $\psi^q\gets\theta^{q,T}-\Delta^q,\quad \mathbf{S}^q\gets(r\cos\psi^q,z,r\sin\psi^q)$
        \EndFor
        \State $\mathcal F\gets\mathsf{BSpline}_{\varepsilon}(\mathsf{Extend}(\mathsf{Close}_{\ell}(\mathbf{S}^w,\mathbf{S}^b)))$
        \State $B\gets\mathsf{SelectValid}_{\varepsilon}(\mathsf{Partition}_{\varepsilon}(E,\mathcal F),\mathbf{S}^w,\mathbf{S}^b)$
    \end{algorithmic}
\end{algorithm}

\noindent\emph{Symbols in Algorithm~\ref{alg:blade-generation}.}
Curves $\mathbf h,\mathbf s,\mathbf i,\mathbf o$ bound the meridional domain.
$N,T$ specify grid size and iteration count.
$E,\ell,\varepsilon$ specify the envelope, closure layout and CAD tolerance, and $\bot$ denotes failure.

The prototype is repeated $N_b$ times about the shaft at pitch $2\pi/N_b$ and phase $\theta_0$.
Meridional profiles define revolved supports with prescribed thickness and shaft-bore geometry.
Optional components include splitters, a casing with clearance, and a volute sized from outlet geometry and operating conditions.
Solid and assembly checks return the model or failure diagnostics to the agent.

\subsection{Inverse Parameter Recovery}
\label{sec:inverse-method}
We recover editable rotor blade parameters from STEP geometry or a STL mesh. After aligning the shaft, we identify both blade sides and recover the meridional geometry of Sec.~\ref{pde_method}. The hub and shroud profiles $C_h,C_s$ and the observed leading edge give the blade starts $t_h,t_s$ and the inlet curve $\mathbf{i}(v)$.

Both formats fit unwrapped circumferential angles on each blade side $q$. Let $\varepsilon_i^q=\widetilde{\theta}^q(u_i,v_i)-\widetilde{\theta}_{i,\mathrm{obs}}^q$ be the angular residual at sample $i$, and $r_i$ its distance to the shaft axis. The fitting losses are
\begin{align}
    \mathcal L_{\mathrm{STEP}}^q&=\sum_i(\varepsilon_i^q)^2 ,
    \label{eq:inverse-boundary-loss}\\
    \mathcal L_{\mathrm{STL}}^q&=
    \frac{|\mathcal S_{\mathrm{int}}|}{|\mathcal S_{\mathrm{int}}|+|\mathcal S_{\mathrm{wall}}|}
    \sum_{\mathcal S}\frac{\sum_{i\in\mathcal S}(r_i\varepsilon_i^q)^2}{|\mathcal S|} .
    \label{eq:inverse-mesh-loss}
\end{align}
For STEP, cubic fits to sampled blade traces give the boundary angle laws.
For STL, we jointly fit the boundary laws and $\kappa^q$ using interpolated PDE angles at interior triangle centers $\mathcal S_{\mathrm{int}}$ and boundary-law angles at resampled hub and shroud points $\mathcal S_{\mathrm{wall}}$. 

The angular change along the chord gives the wrap $\phi_e^q$. For $\lambda^q>0$, the fitted endpoint slopes in Eq.~\eqref{eq:hub-law} give
\begin{equation}
    \beta_e^q=\operatorname{atan2}\bigl(
       \lambda^q,(f_e^q)'(u)-(1-\lambda^q)\phi_e^q\bigr).
    \label{eq:inverse-angle}
\end{equation}
At $u=0$, this gives the inlet control $\beta_{\mathrm{in},e}^q$. At $u=1$, it gives the outlet control $\beta_{\mathrm{out},e}^q$.
The spanwise phase change gives $\delta^q$. At the hub inlet, subtracting the suction-side phase from the pressure-side phase gives $\Delta^b-\Delta^w$.
These controls return to the forward generator for reconstruction and editing.

\subsection{Representation-Grounded Agent}
\label{sec:agent}
The agent accesses the geometric representation, optional parameter correction, and CAD operators through domain tools.
It follows a ReAct loop~\cite{yao2023react}.
At round $t$, it selects tool calls $\mathcal A_t$ from the request $q$, recent conversation and current-turn tool history $H_t$, and available observations $O_t$,
\begin{equation}
    \mathcal A_t=A(q,H_t,O_t),\quad
    O_{t+1}=\operatorname{Execute}(\mathcal A_t).
    \label{eq:agent-update}
\end{equation}
Each call specifies a tool.
Observations enter the next round until a final response or the round limit.

Domain actions cover duty-based initialization, generation and editing, inverse recovery, component construction, and inspection.
System instructions distinguish new operating conditions from edits to existing geometry.
Generation arguments control blade count, wrap, angle field, support dimensions, and splitter placement. Inspection reads properties or reports.

Generation returns file paths, effective configuration, dimensions, part counts, and validity.
Geometric tools report inadmissible configurations, crossing surfaces, invalid solids, and splitter interference.
These observations guide subsequent decisions. 
Instructions distinguish correctable argument errors from native CAD failures. 
Reports of native failures preserve the requested geometry.

\paragraph{State inheritance and revision}
A geometry edit supplies a source state $\mathcal D^{\mathrm{src}}$ through its parameter and a partial update $\Delta\mathcal D$,
\begin{equation}
    \mathcal D^{\mathrm{new}}=
    \operatorname{Merge}(\mathcal D_{\mathrm{src}},\Delta\mathcal D).
    \label{eq:state-merge}
\end{equation}
The tool layer applies the requested overridden value and preserves the original values of others.
Construction determines the geometric consequences of each update.
Each generation saves new CAD and parameter versions for subsequent edits.
The designer can use an earlier parameter version as the starting point for a different edit.

\section{Experiments}
\label{sec:experiments}

We evaluate generation quality, rotor construction, inverse editing, angle-field control, CFD Assessment, and agent cost.
The geometric comparison covers the illustrated machinery types. Construction and inverse editing cover the supported structural configurations.

\begin{figure*}[t]
    \centering
    \includegraphics[width=0.95\linewidth]{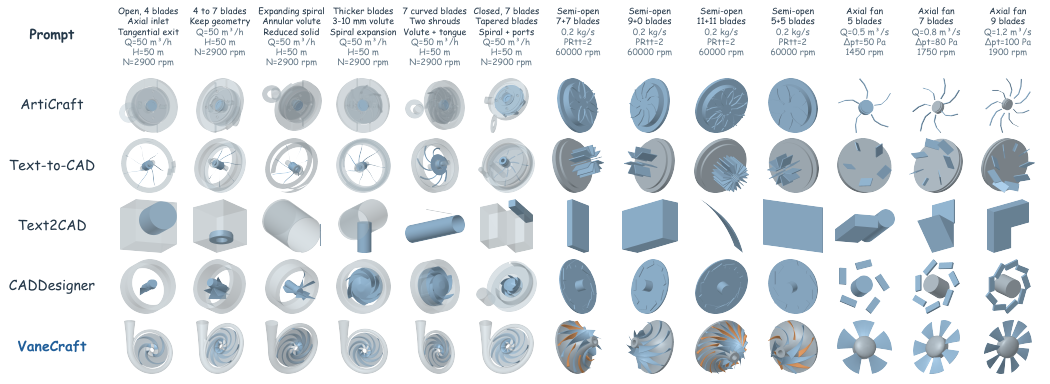}
    \caption{Qualitative comparison of VaneCraft with other generation methods. VaneCraft is able to preserve the periodicity of blades, support structures, and shells, while other methods cannot understand the structure of the fluid rotor and can only generate discrete or simplified primitives.}
    \label{fig:generation}
\end{figure*}

\begin{table}[t]
    \centering
    \small
    \caption{(a) Quantitative comparison of generation methods. (b) Ablation of geometric parameter correction and the surface loss.}
    \label{tab:geometry-comparison}

    \setlength{\tabcolsep}{2pt}
    \renewcommand{\arraystretch}{1.0}
    \begin{tabular*}{0.85\linewidth}{@{\extracolsep{\fill}}clrrr@{}}
        \toprule
        & Method & IoU$\uparrow$ & CD$\downarrow$ & HD$\downarrow$\\
        \midrule
        \multirow{5}{*}{(a)} & ArtiCraft & 0.1155 & 0.0133 & 0.2826\\
        & Text-to-CAD & 0.1102 & 0.0124 & 0.2615\\
        & Text2CAD & 0.0617 & 0.0285 & 0.3518\\
        & CADDesigner & 0.0936 & 0.0136 & 0.2847\\
        & \textbf{VaneCraft} & \textbf{0.1945} & \textbf{0.0093} & \textbf{0.2437}\\
        \midrule
        \multirow{3}{*}{(b)} & w/o GPC & 0.20101 & 0.0053949 & 0.23566\\
        & w/o surface loss & 0.21056 & 0.0050311 & 0.23338\\
        & with GPC & \textbf{0.21332} & \textbf{0.0050211} & \textbf{0.23201}\\
        \bottomrule
    \end{tabular*}
\end{table}

\subsection{Implementation}
\label{sec:exp_setup}
VaneCraft uses Python 3.13. We develop the framework and perform experimental validation on a workstation with an Intel Core Ultra 9 285K CPU and an NVIDIA GeForce RTX 4090 GPU, and on a server with two Intel Xeon 6740E CPUs and two NVIDIA H200 GPUs.
Construction checks validity and part structure. Editing checks requested and retained parameters. Reconstruction compares volume, parts, and sampled distances in the original units.

\paragraph{Datasets}
We train the unified correction model using geometry extracted from SHIFT-Pump~\cite{shift_pump_2026}, ParaturboCAD~\cite{massoudi2024paraturbodata}, and NASA aerodynamic dataset~\cite{sanguineti2022rotor37} for closed, semi-open, and fully open rotors, respectively.
These datasets also provide reference geometries for the geometric comparisons.
We train joint GPC from scratch using zero-correction PDE meshes and their meridional curves as inputs. 
Reference curves and blade surfaces supervise the model.
Forward evaluation supplies GPC with the mesh and meridional curves generated from initialized controls by the PDE representation, then rebuilds geometry from the corrected controls.

\paragraph{Parameter settings}
For inverse recovery, we set the angle-control strength $\lambda^q=1$ on both blade sides.
GPC uses one mesh block with two attention heads and a SiLU decoder with radial and axial heads. The loss weights are $\eta=0.02$ and $\gamma=0.25$.
Auxiliary weights $(w_h,w_s,w_i,w_b)$ are $(0.5,0.5,1,1)$ for closed and semi-open configurations and $(1,1,1,0)$ for full-open configurations.
The geometry penalty weights are $(\mu_1,\mu_2,\mu_3)=(10,0,0)$ in closed, semi-open, and full-open order.
The reference diameter $D$ is the outlet diameter for radial configurations and inlet diameter for full-open configurations.

\begin{table}[t]
    \centering
    \small
    \caption{Generation includes GPC, construction, and CAD export. Inverse reconstruction starts from the generated CAD and includes recovery, rebuilding, export, and geometric verification.}
    \label{tab:family-construction}
    \setlength{\tabcolsep}{2pt}
    \begin{tabular*}{0.85\linewidth}{@{\extracolsep{\fill}}lcccr@{}}
        \toprule
        Configuration & Blades & Parts & Gen. (s) & Inver. (s)\\
        \midrule
        Closed  & $10+0$ & 12 & 5.525 & 18.134\\
        Semi-open$+$splitter  & $7+7$ & 15 & 9.966 & 27.633\\
        Semi-open  & $7+0$ & 8 & 5.857 & 16.554\\
        Open & $16+0$ & 17 & 5.156 & 19.426\\
        Fan & $5+0$ & 6 & 4.445 & 17.378\\
        \bottomrule
    \end{tabular*}
\end{table}

\begin{figure}[t]
    \centering
    \includegraphics[width=0.95\linewidth]{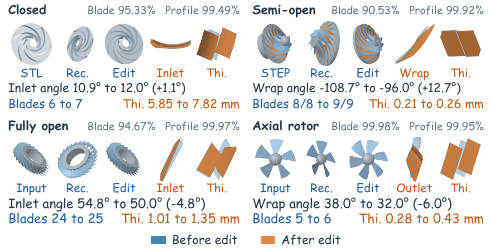}
    \caption{Inverse recovery and continued editing. Blade and Profile give the similarity of the blade surfaces and the meridional curves, respectively. }
    \label{fig:inverse-editing}
\end{figure}

\subsection{Qualitative Evaluation}
\label{sec:exp_generation}
As shown in Fig.~\ref{fig:generation}, the qualitative experiments compare VaneCraft with artiraft \cite{zhou2026articraft}, text-to-cad \cite{texttocad2026}, Text2CAD \cite{khan2024text2cad}, and CADDesigner \cite{fan2026caddesigner}.
VaneCraft adapts the construction to different supports and layouts while preserving the intended periodic pattern and component placement. 
Other comparison methods fail to understand the design features required by the language description, and the generated structures do not meet the design requirements. 
Text2CAD, trained on a basic sequence dataset, is only suitable for generating primitives. ArtiCraft and Text-to-CAD utilize agent skills, achieving similar structures but lacking detailed implementation. 
CADDesigner understands requirements from image rendering, but it cannot understand angular information.

\subsection{Quantitative Evaluation}
\label{sec:exp_quantitative}
Table~\ref{tab:geometry-comparison}(a) compares generated rotor structures with the reference geometries.
Each model is scaled to fit a unit cube and aligned by rotation and translation using the iterative closest point (ICP) algorithm. 
We compute surface-voxel IoU, symmetric squared Chamfer distance (CD), and sampled Hausdorff distance (HD), averaging within rotor types and then equally across types.
CAD generation and inverse reconstruction times for the supported rotor types are reported in Table~\ref{tab:family-construction}.
VaneCraft achieves the highest IoU and the lowest CD and HD.

\begin{table}[t]
    \centering
    \footnotesize
    \setlength{\tabcolsep}{1.5pt}
    \caption{Ablation of parameter support (On) and unassisted LLM parameterization (Off). Acc. is the fraction of matched generation, editing, and recovery tasks passing all execution and parameter checks. Geom. reports generation geometry compliance over eligible generation tasks.}
    \label{tab:agent-workflow-cost}
    \renewcommand{\arraystretch}{1.08}
    \begin{tabular*}{0.85\linewidth}{@{\extracolsep{\fill}}lcccccl@{}}
        \toprule
        & \multicolumn{2}{c}{Total time (s)} & \multicolumn{2}{c}{Acc.} & \multicolumn{2}{c}{Geom.}\\
        \cmidrule(lr){2-3}\cmidrule(lr){4-5}\cmidrule(lr){6-7}
        Model & Off & On & Off & On & Off & On\\
        \midrule
        qwen3.8-max & 562.3 & 385.9 & 0.316 & 0.974 & 0.133 & 1.000\\
        qwen3.8-27b & 428.7 & 267.7 & 0.455 & 1.000 & 0.200 & 1.000\\
        deepseek-v4.1-flash & 148.9 & 96.9 & 0.480 & 1.000 & 0.206 & 1.000\\
        gpt-6.1-sol & 122.2 & 110.7 & 1.000 & 1.000 & 1.000 & 1.000\\
        claude-opus-5-5 & 128.4 & 127.5 & 1.000 & 1.000 & 1.000 & 1.000\\
        \bottomrule
    \end{tabular*}
\end{table}

\subsection{Ablation Study}
\label{sec:exp_robustness_efficiency}
\paragraph{Parameter correction}
Table~\ref{tab:geometry-comparison}(b) compares template initialization, GPC without surface loss, and full GPC.
The hub and shroud profiles describe the blade in the axial--radial plane. Blades can match in this plane.
The surface loss measures this difference in 3D by comparing sampled points on the reconstructed and reference blades.
Its gradients pass through the PDE reconstruction to the predicted control offsets.
GPC without surface loss improves all metrics over template initialization.
Adding surface supervision gives the highest IoU and the lowest CD and HD by a smaller margin, so correcting the meridional controls accounts for most of the gain.

\paragraph{Parameter support}
Table~\ref{tab:agent-workflow-cost} compares engineering parameter support (On) with unassisted LLM parameterization (Off).
On supplies sizing tools, presets, and reference values.
This support provides initial parameter values and reduces the need to infer them directly from language.
On reduces execution time across backbones and improves task pass rates and geometry compliance for the Qwen and DeepSeek.
Parameter support thus improves efficiency across backbones, with the largest reliability gains for models with lower unassisted pass rates.

\begin{table}[t]
    \centering
    \caption{Angle-field control. Blade fit RMSE, CAD build time, and angular deviations from spanwise and chordwise interpolation.}
    \label{tab:anisotropy}
    \small
    \setlength{\tabcolsep}{2pt}
    \renewcommand{\arraystretch}{1.02}
    \begin{tabular*}{0.85\linewidth}{@{\extracolsep{\fill}}lrrrr@{}}
        \toprule
        Field & \shortstack{RMSE\\(mm)} & \shortstack{Time\\(s)} & $E_{\mathrm{span}}$ ($^\circ$) & $E_{\mathrm{chord}}$ ($^\circ$)\\
        \midrule
        Interp.   & 1.071 & 3.05 & 0.0000 & 9.7363\\
        $\kappa=0.1$       & 15.470 & 4.97 & 9.4589 & 0.0000\\
        $\kappa=0.25$      & 13.961 & 4.74 & 8.6695 & 0.0371\\
        $\kappa=0.5$       & 11.373 & 4.35 & 7.2200 & 0.8363\\
        $\kappa=1$         & 7.001 & 3.95 & 4.5239 & 3.8495\\
        $\kappa=2$         & 2.978 & 3.73 & 1.8721 & 7.3014\\
        $\kappa=10$        & \textbf{1.039} & 3.00 & 0.1055 & 9.6102\\
        $\kappa=30$        & 1.064 & 2.96 & 0.0123 & 9.7221\\
        \bottomrule
    \end{tabular*}
\end{table}

\subsection{Inverse and Editing Reconstruction}
\label{sec:exp_editing}
VaneCraft converts existing STEP and STL rotor models into editable parameters for continued geometric design.
It aligns the shaft, identifies both blade sides, and recovers the meridional profiles and boundary angle laws.
These parameters place the recovered geometry in the same representation used for forward generation.
Fig.~\ref{fig:inverse-editing} shows that the reconstructed blade surfaces and meridional curves closely match the inputs.
Starting from this reconstruction, we change the blade count and the inlet angle or wrap, then adjust local thickness through the angular offset of the back surface.
Each edit updates the parameter state saved in the preceding step.
The PDE regenerates the blade surfaces from the updated controls, and we import each resulting CAD assembly to check solid validity.
Inverse recovery makes existing models reusable for further design changes.

\begin{figure}[t]
    \centering
    \includegraphics[width=0.95\linewidth]{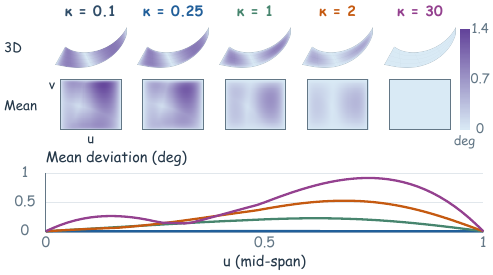}
    \caption{Angle-field control. The 3D views show absolute deviations from spanwise interpolation on a representative work surface $\mathbf{S}^w$. The heatmaps average these deviations across configurations and blade sides. The curves show category-averaged mid-span deviations from chordwise interpolation.}
    \label{fig:angle-field-cd}
\end{figure}

\begin{table}[t]
\centering
\footnotesize
\caption{CFD results at the design operating points. $H_d$ and $H_c$ denote target and calculated head. $\delta_H=|H_c-H_d|$ is the absolute head error in metres. $\eta_h$ is hydraulic efficiency.}
\label{tab:cfd-summary}
\setlength{\tabcolsep}{1.5pt}
\begin{tabular*}{0.85\linewidth}{@{\extracolsep{\fill}}lrrrrr@{}}
\toprule
Case & P1-T & P2-T & P3-R & P4-T & P4-R\\
\midrule
$n$ (rpm) & 5800 & 2900 & 1450 & 1450 & 1450\\
$Q$ (m$^3$/h) & 40 & 25 & 400 & 6.3 & 6.3\\
$H_d$ (m) & 40 & 50 & 35 & 5 & 5\\
$n_q/n_s$ & 38/140 & 13/47 & 38/122 & 18/66 & 18/66\\
$H_c$ (m) & 40.05 & 47.45 & 35.42 & 4.46 & 4.42\\
$\delta_H$ (m) & 0.05 & 2.55 & 0.42 & 0.54 & 0.58\\
$\eta_h$ (\%) & 80.9 & 79.90 & 81.23 & 75.18 & 74.82\\
\bottomrule
\end{tabular*}
\end{table}

\subsection{Angle-Field Control}
\label{sec:PDE-based surface}
The anisotropy parameter $\kappa$ controls the balance between chordwise and spanwise propagation of the boundary angle laws.
Spanwise interpolation blends the hub and shroud angles as $(1-v)\theta_h^q(u)+v\theta_s^q(u)$. At $\kappa=1$, chordwise and spanwise coupling have equal weights. Larger $\kappa$ gives the hub and shroud laws more influence on the interior.
All fields share the meridional geometry, boundary angles, and evaluation points within each case.
Fit RMSE compares blade points at the same axial position and radius.
As $\kappa$ increases, the interior angular RMSE from spanwise interpolation ($E_{\mathrm{span}}$) falls, as shown in Table~\ref{tab:anisotropy}.
At mid-span, the angular RMSE from linear interpolation between inlet and outlet angles ($E_{\mathrm{chord}}$) rises.
Fig.~\ref{fig:angle-field-cd} shows this shift under fixed boundaries.
These results show that VaneCraft can adjust the interior blade shape through $\kappa$ while preserving the prescribed boundaries.

\subsection{CFD Assessment of Generated Designs}
\label{sec:exp_cfd}
\begin{figure}[t]
    \centering
    \IfFileExists{fig/pump_panel.pdf}{\includegraphics[width=0.95\linewidth]{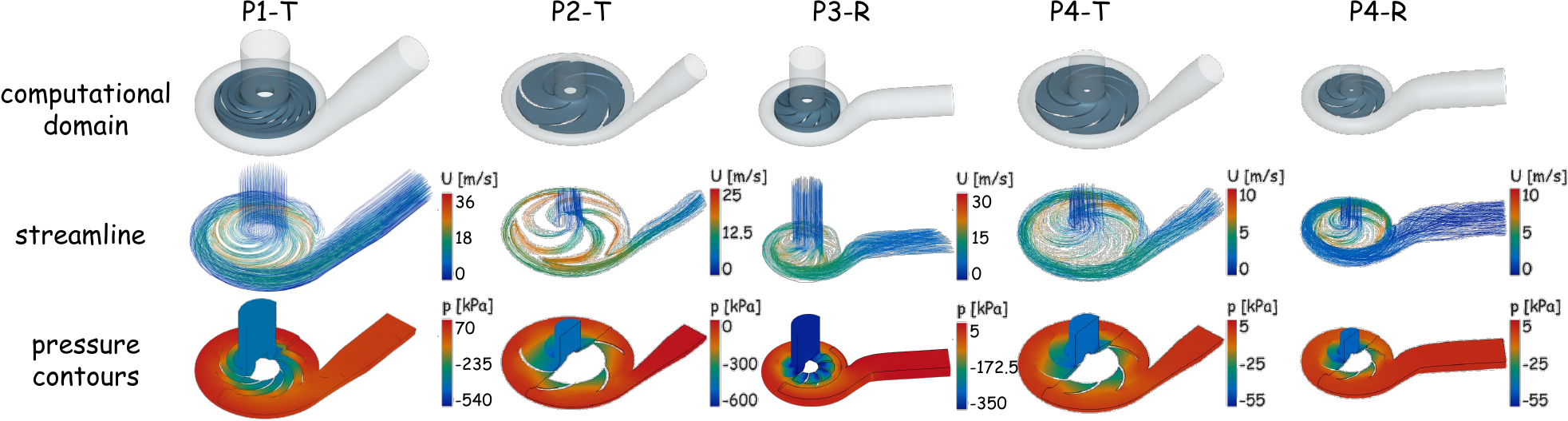}}{\fbox{\parbox[c][35mm][c]{\linewidth}{\centering Pump fluid-domain sections and CFD streamlines}}}
    \caption{CFD streamlines for the closed cases. Streamlines are traced with relative velocity in the rotating region and absolute velocity in the stationary region.}
    \label{fig:cfd-pump}
\end{figure}

\begin{figure}[t]
    \centering
    \IfFileExists{fig/fan_compressor_group.pdf}{\includegraphics[width=0.95\linewidth]{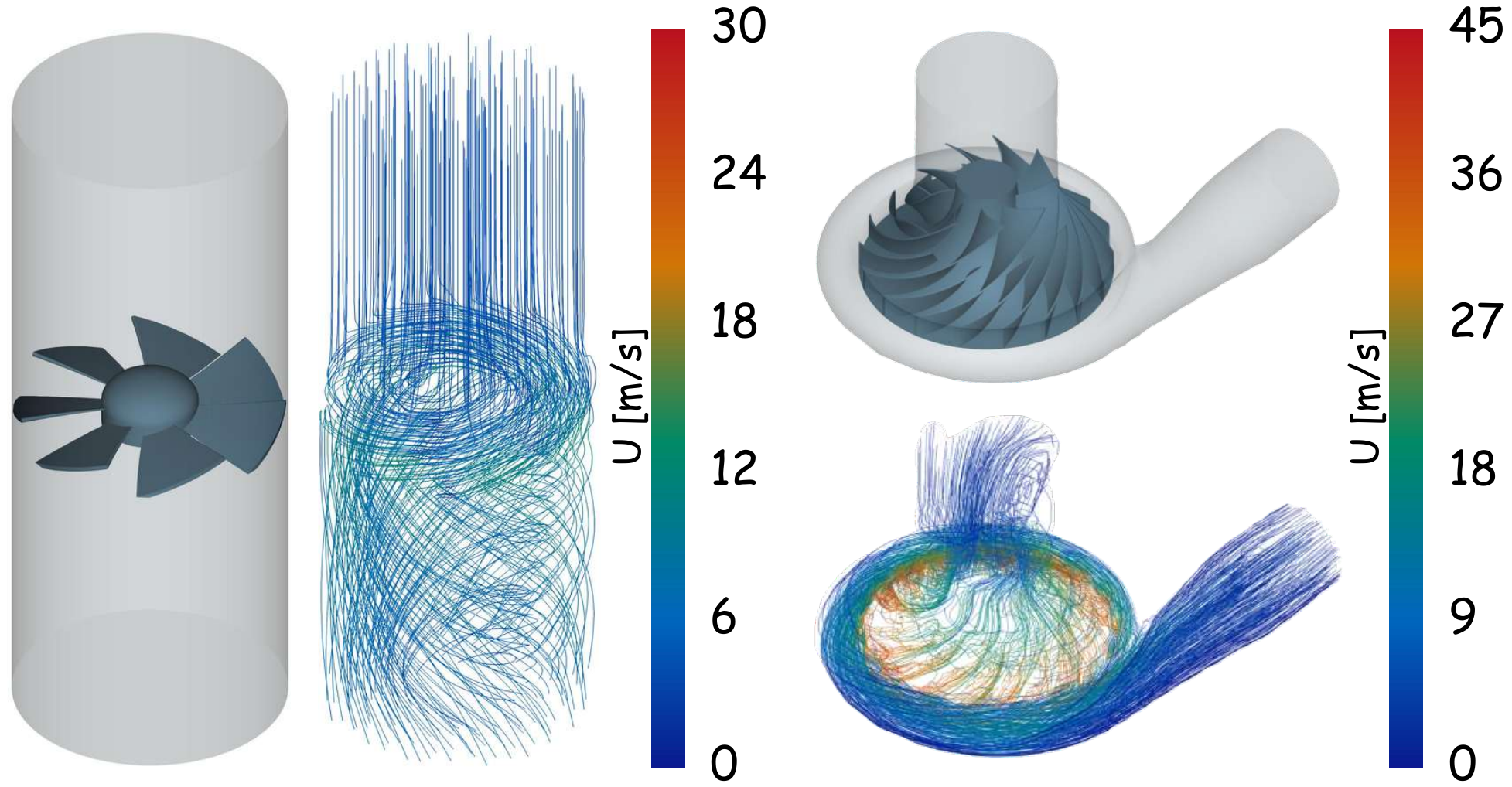}}{\fbox{\parbox[c][30mm][c]{0.95\linewidth}{\centering Fan and compressor flow fields}}}
    \caption{Flow fields of the generated Semi-open and full-open. Streamlines are colored by velocity magnitude.}
    \label{fig:cfd-fan-compressor}
\end{figure}

We use CFD to assess the performance and flow patterns of the rotors generated by VaneCraft.
The simulations used OpenFOAM v2606 with steady MRF, $k$--$\omega$ SST, and adaptive wall treatment~\cite{huang2019openfoam,petit2013pump}. Each case used an independent mesh, a velocity inlet, and a pressure outlet. 
Convection used upwind for the first 200 iterations and limitedLinearV thereafter.
Head was computed from the difference in flow-weighted mean total pressure between the inlet and the outlet, and hydraulic efficiency as $\eta_h=\rho gQH/P_{\mathrm{rotor}}$, where $P_{\mathrm{rotor}}$ is the power transferred to the fluid by the rotor.
As shown in Table~\ref{tab:cfd-summary}, the designs achieve favorable hydraulic efficiencies. Fig.~\ref{fig:cfd-pump} shows streamlines through the rotor and casing. Aerodynamic simulations on the generated flow domains show downstream swirl for the fully open rotor and radial discharge for the semi-open rotor in Fig.~\ref{fig:cfd-fan-compressor}.
CFD predicts performance close to the specified design targets for the rotors generated by VaneCraft.

\section{Conclusion}
\label{sec:conclusion}
VaneCraft unifies rotor generation, inverse recovery, and continued editing through a controllable analytical representation.
Explicit geometric parameters provide a consistent basis for agents to perform CAD modeling.
The PDE construction and geometric parameter correction combine control over blade shape with fitting to reference geometry.
Experiments demonstrate controllable interior geometry and parameter editing, with the highest surface IoU and lowest Chamfer distance in the rotor comparison.
Inverse recovery extends these capabilities to existing CAD models and preserves their editability across successive revisions.
The unified representation helps users with limited CAD experience generate, reuse, and modify different types of rotors.
\par

{
    \small
    \bibliographystyle{ieeenat_fullname}
    \bibliography{main}
}

\clearpage
\setcounter{page}{1}
\maketitlesupplementary
\renewcommand{\thesection}{S\arabic{section}}
\setcounter{section}{0}

\section{Initial Parameters and Meridional Geometry}
\label{app:notation}
\begin{figure}[ht!]
    \centering
    \includegraphics[width=0.95\linewidth]{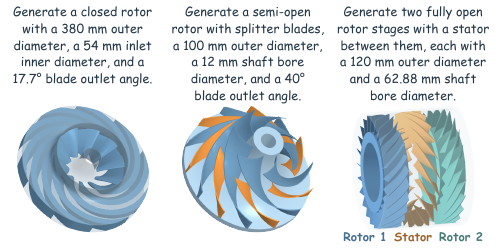}
    \caption{Design instructions and the resulting rotor models.}
    \label{fig:direct-dimensions}                                                             
\end{figure}

The state $\mathcal D=(k,\mathbf p,\boldsymbol\xi)$ stores the rotor category, rotor controls, and component parameters. A forward design request specifies the rotor category and can include explicit controls. A geometric preset supplies initial values when generation starts from a template. Users can also generate a rotor by specifying its principal dimensions directly, as shown in Fig.~\ref{fig:direct-dimensions}.

\begin{figure}[t]
    \centering
    \includegraphics[width=0.95\linewidth]{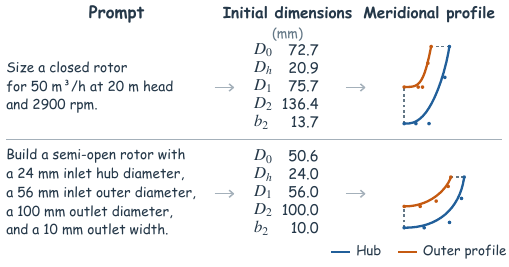}
    \caption{Initial dimensions and meridional profile derived from operating conditions. Dots mark profile control points.}
    \label{fig:initial-sizing}
\end{figure}

\paragraph{Sizing from design requirements}
The work side $w$ is the pressure side, and the back side $b$ is the suction side. 
Supplied principal dimensions can directly initialize the hub profile and the shroud or outer blade profile. 
When the design specifies operating conditions, we first estimate the dimensions as shown in Fig.~\ref{fig:initial-sizing}.

\begin{figure*}[htbp!]
    \centering
    \includegraphics[width=0.95\linewidth]{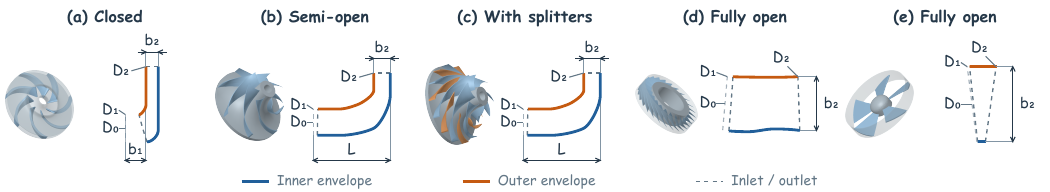}
    \caption{Rotor assemblies (left) and meridional profiles (right). Axial lengths are marked on the shroud in (a) as $b_1$ and on the hub in (b,c) as $L$. Sizing uses the pressure ratio in (d) and the pressure rise in (e).}
    \label{fig:machine-families}
\end{figure*}

The initial meridional inlet and outlet outer diameters are $D_1$ and $D_2$. The area-equivalent inlet diameter $D_0$ matches a circle to the geometric area of the inlet annulus projected normal to the shaft, before correction. 
In this work, the closed and semi-open rotors are radial, and fully open rotor configurations are axial.
At the outlet, $b_2$ denotes the axial passage width for radial rotors and the radial blade height for axial rotors.

The closed rotor requires volume flow $Q$ in $\mathrm{m^3/s}$, head $H$ in metres, and speed $n$ in rpm. 
Set $S=(Q/n)^{1/3}$ and specific speed $n_s=3.65n\sqrt Q/H^{3/4}$. The dimensions are
\begin{equation}
\begin{aligned}
D_0&=k_0S, \quad D_1=\sqrt{D_0^2+D_h^2},\\
D_2&=9.35k_D(n_s/100)^{-1/2}S,\\
b_2&=0.64k_b(n_s/100)^{5/6}S ,
\end{aligned}
\end{equation}
where $D_h$ is the hub inlet diameter. 
The coefficients $k_0$, $k_D$, and $k_b$ correspond to inlet diameter, outlet diameter, and outlet width, respectively. 
Users can specify each coefficient directly or use its default value, obtained by linear interpolation between two adjacent table points at the specific speed $n_s$.

For semi-open rotors, the user supplies mass flow $\dot m$, rotational speed $n$, total pressure ratio $\Pi$, inlet total pressure $p_{t1}$, and inlet total temperature $T_{t1}$.
At the inlet, the open-area factor $a_1$ defaults to $0.96$, and the ratio $\nu_1$ of hub diameter to outer diameter defaults to $0.40$.
Inlet continuity gives the area-equivalent diameter $D_0=\sqrt{4\dot m/(\pi a_1\rho_1 C_{m1})}$ and the inlet outer diameter $D_1=D_0/\sqrt{1-\nu_1^2}$, where $\rho_1$ and $C_{m1}$ are the inlet static density and meridional speed.
The remaining adjustable defaults are compression efficiency $\eta_{\mathrm{is}}=0.80$, slip factor $\sigma=0.90$, outlet flow coefficient $\varphi=0.20$, outlet blade angle $\beta_2=50^\circ$, and outlet open-area factor $a_2=0.94$.
The work per unit mass $W$ is 
\begin{equation}
    W=c_p T_{t1}[\Pi^{(\gamma_g-1)/\gamma_g}-1]/\eta_{\mathrm{is}},
\end{equation} 
where the gas properties $c_p$ and $\gamma_g$ are the specific heat at constant pressure and the ratio of specific heats.
The outlet blade and meridional speeds are $U_2=\sqrt{W/(\sigma-\varphi\cot\beta_2)}$ and $C_{m2}=\varphi U_2$.
The outlet velocity and total state give the static pressure $p_2$ and temperature $T_2$ for $\rho_2=p_2/(RT_2)$, where $R$ is the gas constant.
With angular speed $\omega=2\pi n/60$, the outlet dimensions are $D_2=2U_2/\omega$ and $b_2=\dot m/(\pi D_2a_2\rho_2 C_{m2})$.

Fully open requires rotational speed $n$ and allow adjustment of the hub-to-outer diameter ratio $\nu$ and inlet open-area factor $a$. The inlet and outlet total pressures $p_{t1}$ and $p_{t2}$ give the ratio $\Pi=p_{t2}/p_{t1}$ and rise $\Delta p_t=p_{t2}-p_{t1}$. $\rho$ is the inlet static density, $C_a$ the inlet axial speed, and $W$ the specific work computed from $\Pi$ as above.
For $(\dot m,\Pi)$, the defaults are $\nu=0.71$, $a=0.96$, work coefficient $\psi_w=0.32$, $p_{t1}=101.325\,\mathrm{kPa}$, and $T_{t1}=298.15\,\mathrm{K}$.
\begin{equation}
\label{eq:open-ratio-diameters}
\begin{aligned}
D_{\mathrm{flow}} &= \sqrt{\frac{4\dot m}{\pi a\rho C_a(1-\nu^2)}},\\
D_{\mathrm{load}} &= \frac{2}{\omega}\sqrt{\frac{2W}{\psi_w(1+\nu^2)}}.
\end{aligned}
\end{equation}
For $(Q,\Delta p_t)$, the defaults are $\nu=0.40$, $a=0.97$, flow coefficient $\varphi=0.22$, and pressure coefficient $\psi_p=0.08$.
\begin{equation}
\label{eq:open-rise-diameters}
D_{\mathrm{flow}} = \sqrt[3]{\frac{8Q}{\pi a\varphi\omega(1-\nu^2)}}, \quad
D_{\mathrm{load}} = \frac{2}{\omega}\sqrt{\frac{2\Delta p_t}{\rho\psi_p}}. 
\end{equation}

Automatic sizing selects the larger candidate for $D_2$. With a specified pressure rise, the user can also supply $D_2$ directly. 
\begin{equation}
\label{eq:open-annulus-dimensions}
\begin{alignedat}{2}
D_2 &= \max(D_{\mathrm{flow}},D_{\mathrm{load}}), &\quad D_1 &= D_2,\\
D_0 &= D_1\sqrt{1-\nu^2}, & b_2 &= \tfrac{D_2}{2}(1-\nu).
\end{alignedat}
\end{equation}
Changing the dimensions preserves the component structure of the CAD model and affects the rotor performance predicted by CFD.

\paragraph{From dimensions to controls}
Profile and blade-angle controls define the initial PDE blade for optional GPC.

Radial profiles use separate hub and shroud control-point ratios. 
Closed rotors use specific speed for the outlet angle and a preset wrap range. 
Semi-open rotors use design coefficients for both.

For fully open rotors, velocity triangles give relative flow angles $\beta_i$. The PDE angle $\alpha_i$ sets the end slope $\cot\alpha_i=(\ell_z/r)\cot\beta_i$, with axial chord $\ell_z$ and section radius $r$. When sizing starts from $\Pi$, the wrap in radians is the mean end slope. For a specified pressure rise $\Delta p_t$, the wrap angle in radians equals the circumferential distance from the leading edge to the trailing edge divided by the section radius $r$.

The hub and shroud profiles $C_h,C_s$ use the Bezier representation in Eq.~\eqref{eq:bezier}.
Users can set $t_h$ and $t_s$ independently or use their default values to locate the inlet points $C_h(t_h)$ and $C_s(t_s)$ on the hub and shroud curves.
The retained downstream segments become the blade-domain boundaries.
The leading-edge meridional curve is a quadratic Bezier curve through the two endpoint positions.
Its middle control point is the endpoint midpoint displaced radially by $\delta_r$.
The trailing-edge curve joins the downstream profile endpoints.

Each boundary is resampled by linear interpolation at uniformly spaced cumulative polyline arc lengths.
The Coons corner correction in Eq.~\eqref{eq:coons} is
\begin{equation}
\begin{split}
    \mathbf{co}(u,v)={}&(1-u)(1-v)\mathbf{h}(0)
                       +u(1-v)\mathbf{h}(1)\\
                     &+(1-u)v\mathbf{s}(0)+uv\mathbf{s}(1).
\end{split}
\end{equation}
The resulting interpolation reproduces the four boundary curves at $u=0,1$ and $v=0,1$.
The forward generator reuses recovered meridional profiles through shape-preserving interpolation of their ordered samples.

\begin{table*}[h]
    \centering
    \small
    \setlength{\tabcolsep}{3pt}
    \caption{Geometric evaluation results from Table~\ref{tab:geometry-comparison} by rotor configuration.}
    \label{tab:geometry-by-construction}
    \begin{tabular*}{0.95\linewidth}{@{\extracolsep{\fill}}l*{9}{r}c@{}}
        \toprule
        & \multicolumn{3}{c}{Closed} & \multicolumn{3}{c}{Semi-open} & \multicolumn{3}{c}{Fully open} & \multirow{2}{*}{\shortstack{Generation \\ time (s)}}\\
        \cmidrule(lr){2-4}\cmidrule(lr){5-7}\cmidrule(lr){8-10}
         Generation method & IoU$\uparrow$ & CD$\downarrow$ & HD$\downarrow$ & IoU$\uparrow$ & CD$\downarrow$ & HD$\downarrow$ & IoU$\uparrow$ & CD$\downarrow$ & HD$\downarrow$ & \\
        \midrule
        ArtiCraft & 0.1684 & 0.0108 & 0.2662 & 0.1226 & 0.0132 & 0.3005 & 0.0556 & 0.0159 & 0.2811 & 770.88 \\
        Text-to-CAD & 0.0550 & 0.0190 & 0.2929 & 0.1597 & 0.0097 & 0.2345 & 0.1160 & 0.0085 & 0.2571 & 7.26\\
        Text2CAD & 0.0551 & 0.0374 & 0.3974 & 0.0606 & 0.0332 & 0.3876 & 0.0694 & 0.0150 & 0.2704 & 20.40 \\
        CADDesigner & 0.1140 & 0.0127 & 0.2843 & 0.0878 & 0.0193 & 0.3328 & 0.0789 & 0.0089 & 0.2369 & 476.63\\
        VaneCraft & 0.2406 & 0.0139 & 0.2747 & 0.2247 & 0.0082 & 0.2336 & 0.1181 & 0.0057 & 0.2228 & 4.23\\
        \bottomrule
    \end{tabular*}
\end{table*}

\section{Boundary Laws }
\label{app:angle-field}
For each anisotropy setting, we solve both blade sides on a $61\times61$ grid with 20,000 Jacobi updates. The sweeps fix the meridional geometry and boundary laws, as shown in Fig.~\ref{fig:machine-families}. 

\paragraph{boundaries}
For the closed dataset, we recover the meridional domain from each STL and fit cubic angle laws to blade-wall seams. 
For the other sets, we derive shared meridional profiles by averaging the inner and outer boundary traces of the two blade sides.

Fit RMSE uses mesh triangle centers, STEP surface samples, or native point-cloud samples projected into the common meridional domain. Fields share the evaluation points in alternating blocks of a $6\times6$ partition with $u,v\in(0.05,0.95)$. 
At matched axial position and radius $r$, an angular residual $\Delta\theta$ gives a distance of $2r|\sin(\Delta\theta/2)|$.

Let $I$ contain the interior grid nodes $(u,v)$ and $J$ the interior chord samples $u$. On each side, the spanwise residual is $e_s(u,v)=\theta(u,v)-[(1-v)\theta(u,0)+v\theta(u,1)]$. The chordwise residual is $e_c(u)=m(u)-[(1-u)m(0)+um(1)]$, where $m(u)=\theta(u,1/2)$ averages the two central rows on an even grid. The angular RMSEs in degrees are
\begin{equation}
\begin{aligned}
 E_{\mathrm{span}}&=\frac{180}{\pi}\sqrt{\frac{1}{|I|}\sum_{(u,v)\in I}e_s(u,v)^2},\\
 E_{\mathrm{chord}}&=\frac{180}{\pi}\sqrt{\frac{1}{|J|}\sum_{u\in J}e_c(u)^2}.
\end{aligned}
\end{equation}

At the inlet ($j=0$) and outlet ($j=1$), the boundary values interpolate between the hub and shroud,
\begin{equation}
    \theta^q(j,v)=(1-v)\theta_h^q(j)+v\theta_s^q(j),
    \quad j\in\{0,1\}.
    \label{eq:edge-interp}
\end{equation}

\begin{figure}[H]
    \centering
    \includegraphics[width=0.95\linewidth]{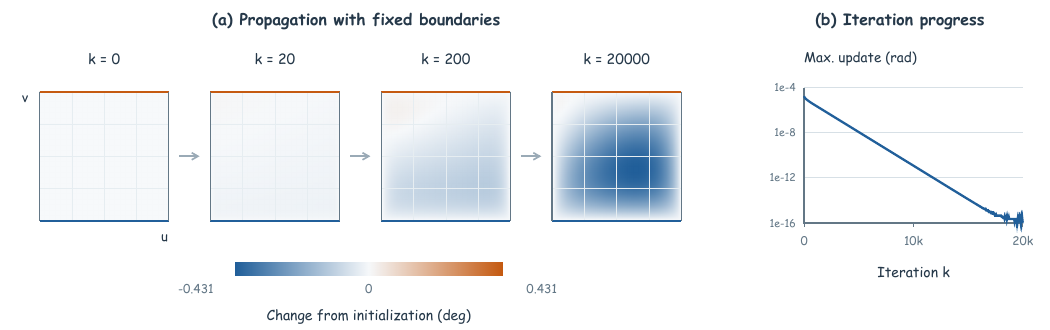}
    \caption{a shows the change in blade angle from the initial spanwise interpolation, and b shows the maximum absolute angular update per iteration.}
    \label{fig:angle-field-e}
\end{figure}

Algorithm~\ref{alg:blade-generation} starts from spanwise interpolation and reapplies the boundary values after each Jacobi update. 
For positive anisotropy, the weights form a convex average, keeping all iterates within the boundary range. 
Fig.~\ref{fig:angle-field-e} shows that the interior angle field evolves and the largest absolute angular change.

\section{Geometric Parameter Correction}
\label{app:unified-correction}
\paragraph{Inputs}
The encoder takes the rotor category $k$, the triangular mesh $\mathcal M$ of a single blade from the initial PDE model, and the corresponding hub, shroud, and inlet curves.
\paragraph{Encoder and decoder}
The signed incidence maps $d_0$ and $d_1$ take nodes to edges and edges to faces, respectively. 
Projected values $V_n,V_e,V_f$ give updates $Y_n,Y_e,Y_f$. 
Each $A$ is local attention for its element type with a meridional bias, and the superscript $\mathrm r$ swaps queries and keys.
\begin{align}
Y_n &= A_n^{\mathrm r}\!\left(d_0^\top A_e(d_0V_n)\right),\label{eq:gpc-node-update}\\
Y_e &= A_e\!\left(d_1^\top A_f(d_1V_e)\right) + d_0 A_n^{\mathrm r}\!\left(d_0^\top A_e(V_e)\right),\label{eq:gpc-edge-update}\\
Y_f &= d_1 A_e^{\mathrm r}\!\left(d_1^\top A_f(V_f)\right).\label{eq:gpc-face-update}
\end{align}
Node values pass through edges and back to nodes, as shown in Fig.~\ref{fig:gpc-overview}(b). 
Face values pass through edges and back to faces. 
Edge updates sum separate paths through faces and nodes. 
The decoder uses pooled updates to predict radial or axial control offsets, as shown in Fig.~\ref{fig:gpc-overview}(a).
We use 70\% of the samples for training, 15\% for validation, and 15\% for testing.

\section{Generation Tasks and Geometric Evaluation}
\label{app:generation-tasks}
\paragraph{Comparison metrics}
Point-to-point iterative closest point (ICP) aligns the generated mesh to the reference using 1,024 area-weighted samples per mesh. 
For aligned sets $A,B$ of 2,048 area-weighted points each, let $d(x,S)=\min_{y\in S}\|x-y\|_2$. 
We compute the squared Chamfer distance and sampled Hausdorff distance in Table~\ref{tab:geometry-by-construction} as
\begin{equation}
\begin{aligned}
\mathrm{CD} &= \frac{1}{|A|}\sum_{a\in A}d(a,B)^2
               +\frac{1}{|B|}\sum_{b\in B}d(b,A)^2,\\
\mathrm{HD} &= \max\!\left\{\max_{a\in A}d(a,B),\;\max_{b\in B}d(b,A)\right\}.
\end{aligned}
\end{equation}

For the CAD generation timings in Table~\ref{tab:geometry-by-construction}, all agents use GPT-6.1-sol as their LLM backbone.
ArtiCraft and CADDesigner take the longest. ArtiCraft writes modeling code for each design, and CADDesigner revises its modeling steps using execution feedback. Text2CAD predicts a sequence of CAD operations from the prompt. Text-to-cad uses predefined CAD skills. Ours has the shortest measured time. 
Starting from initialized parameters, it applies GPC and repeats a blade prototype around the shaft. 
It checks solid validity and part count before STEP export. 
In interactive use, the agent receives the check results and can correct input errors before retrying.

\section{Blade Solids and Splitters}
\label{app:blade-validity}
\begin{figure}[t]
    \centering
    \includegraphics[width=0.95\linewidth]{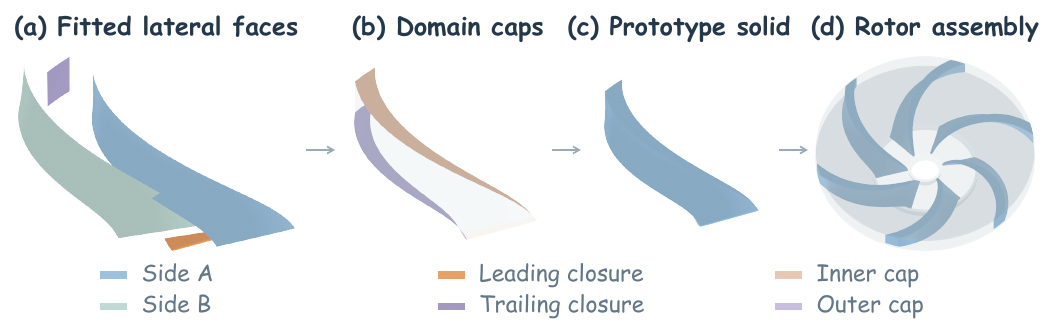}
    \caption{The blade sides are separated for display, and the assembled supports are translucent.}
    \label{fig:blade-solid}
\end{figure}

In Fig.~\ref{fig:blade-solid}, two B-spline surfaces fitted to the blade-side grids join at the leading and trailing edges. 
Checks cover shell closure, side separation, trimming, positive volume, CAD validity, and the expected solid count.
We form each splitter by retaining the portion of a main blade from the selected start to the trailing edge.

\section{Component Operators}
\label{app:pump-components}
\begin{figure}[H]
    \centering
    \includegraphics[width=\columnwidth]{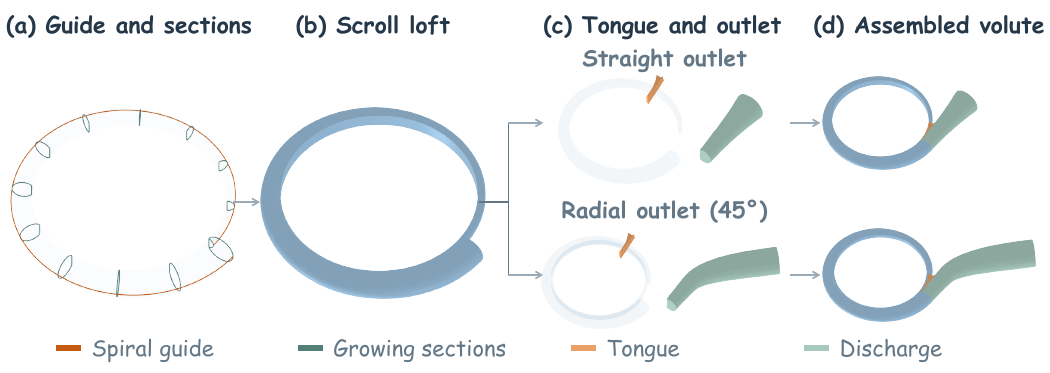}
    \caption{Construction of volutes with straight tangential and radial outlets.}
    \label{fig:volute_construction}
\end{figure}

Rotor dimensions set the initial radius and axial placement of the volute in Fig.~\ref{fig:volute_construction}. Sections grow along an Archimedean guide and are lofted to form the scroll. Both variants use the same tongue connection and keep the outlet centerline in the plane normal to the shaft. 
The straight outlet follows the scroll tangent. The radial outlet turns outward by $45^\circ$ from that tangent after the tongue.

\section{Reconstruction and Editing}
\label{app:reconstruction}
STEP recovery aligns the shaft with the forward $Y$ axis and uses support faces and periodic blade geometry to recover profiles, blade groups, counts, and phase. Open layouts recover the root at the physical hub and the virtual shroud from the outer blade cap. Semi-open recovery with support replacement accepts fused STEP with identifiable periodic blade sides. Closed-rotor STL recovery uses a watertight fluid mesh and length scale. 

\begin{figure}[t]
    \centering
    \includegraphics[width=0.95\linewidth]{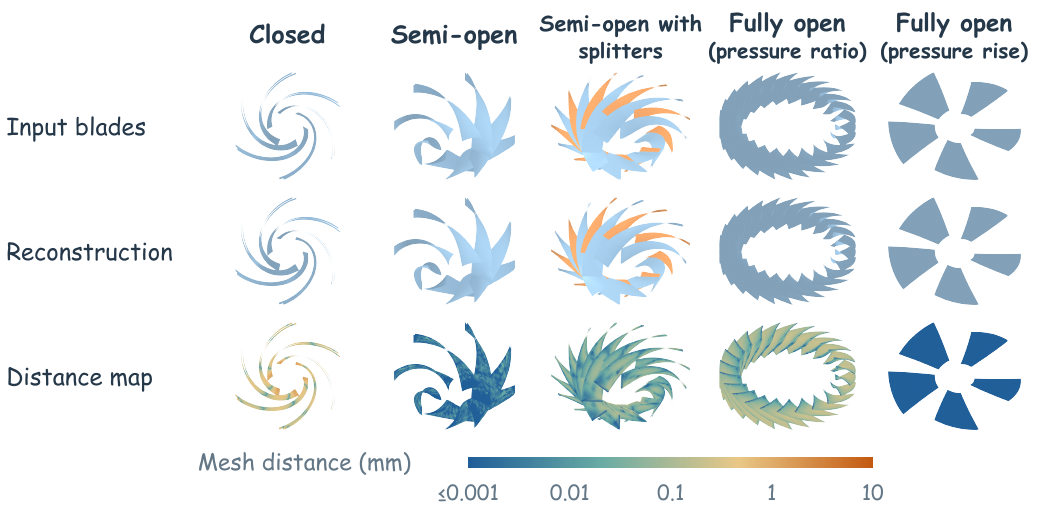}
    \caption{Distance maps show where blades reconstructed from recovered parameters differ from the input blades.}
    \label{fig:inverse-reconstruction}
\end{figure}

\paragraph{Blade fitting}
For observed leading-edge endpoint $p_e$, the profile start is $\widehat t_e=\arg\min_{t\in[0,1]}\|C_e(t)-p_e\|_2^2$.
At the hub inlet, the measured pressure- and suction-side azimuths satisfy $\widetilde{\theta}^w(0,0)-\widetilde{\theta}^b(0,0)=\Delta^b-\Delta^w$ by Eq.~\eqref{eq:map3d}. 
Recovery sets $\Delta^w=0$, so this difference gives the back-surface offset $\Delta^b$ relative to the work surface.
Recovered profiles are stored as ordered samples. 
The STL loss in Eq.~\eqref{eq:inverse-mesh-loss} combines interior and wall angle residuals. 
Interior variation informs the anisotropy fit, and the leading-edge envelope supplies the radial offset where the axial span permits. 
Fig.~\ref{fig:inverse-reconstruction} compares the input blades with those rebuilt from the recovered parameters.

\section{Agent Operations and Evaluation}
\label{app:agent-timing}
We built a dataset of 200 rotor design cases, 40 per configuration, to evaluate how well agents follow design instructions across tasks. 
Each case includes prompts to generate a rotor from operating conditions or dimensions, recover editable parameters from a supplied STEP file, and apply a requested change to the recovered model. 
Settings use the same CAD tools, and On adds sizing tools, presets, and parameter references. 

\begin{figure*}[t]
    \centering
    \includegraphics[width=0.95\linewidth]{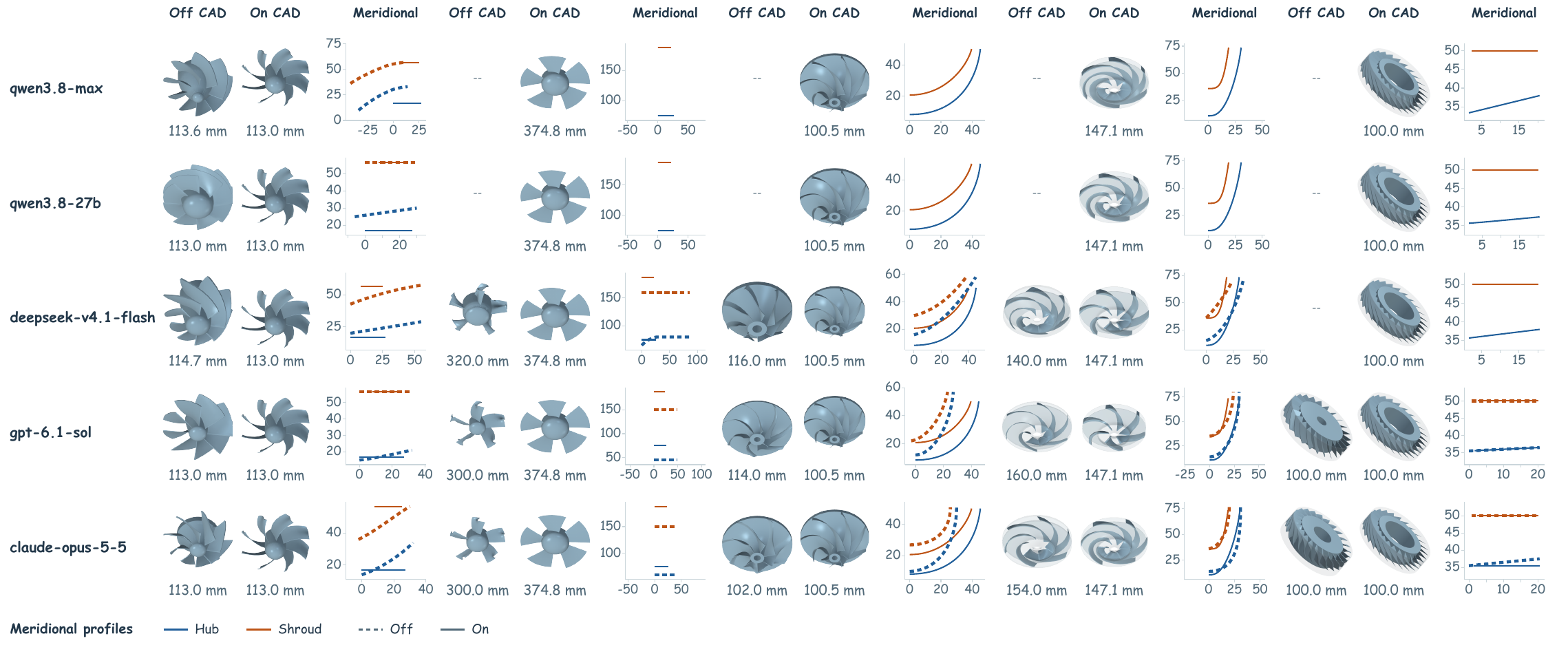}
    \caption{CAD and meridional profiles generated without (Off) and with (On) parameter support.}
    \label{fig:work_atlas}
\end{figure*}

Table~\ref{tab:exp_intent_comparison} compares time and memory use by task, total token use, and both pass rates for each model with and without parameter support. 
Acc. is the number of matched stage tasks passing all execution and parameter checks divided by the number evaluated. 
Geom. is the fraction of eligible generation attempts that produce valid CAD with the requested component counts, periodicity, blade contact with the support, and outer diameter when specified. 
It includes unpaired attempts and counts missing CAD as a failure. 
Fig.~\ref{fig:work_atlas} compares the generated CAD models and overlays their meridional profiles for the same requests with and without parameter support. 
Without parameter support, qwen3.8-27b, qwen3.8-max and deepseek-v4.1-flash reached their output token limits.
Parameter support reduces output token use and shortens forward generation time.

\begin{table*}[t]
  \centering
  \footnotesize
  \renewcommand{\arraystretch}{1.12}
  \caption{Task cost and pass rates with engineering-guided (On) and unassisted (Off) parameterization.}
  \label{tab:exp_intent_comparison}
  \setlength{\tabcolsep}{2.5pt}
  \begin{tabular*}{0.85\linewidth}{@{\extracolsep{\fill}}lllrrrrrrrrr@{}}
    \toprule
    \multirow{2}{*}{Model} & \multirow{2}{*}{Setting} & \multirow{2}{*}{Task} & \multicolumn{3}{c}{Time (s)} & \multirow{2}{*}{\shortstack{RSS\\(MiB)}} & \multicolumn{3}{c}{Tokens (k)} & \multicolumn{2}{c}{Rates}\\
    \cmidrule(lr){4-6}\cmidrule(lr){8-10}\cmidrule(lr){11-12}
    & & & Total & API & CAD & & In & Out & Total & Acc. & Geom.\\
    \midrule
    \multirow{6}{*}{qwen3.8-max} & \multirow{3}{*}{Off} & Forward & 256.7 & 254.2 & 2.5 & 148.5 & \multirow{3}{*}{190.6} & \multirow{3}{*}{15.9} & \multirow{3}{*}{206.5} & \multirow{3}{*}{0.316} & \multirow{3}{*}{0.133}\\
     &  & Modify & 132.4 & 125.0 & 7.3 & 563.2 &  &  &  &  & \\
     &  & Reverse & 173.2 & 145.7 & 27.4 & 564.9 &  &  &  &  & \\
    \cmidrule(l){2-12}
     & \multirow{3}{*}{On} & Forward & 82.5 & 74.9 & 7.4 & 449.4 & \multirow{3}{*}{217.5} & \multirow{3}{*}{7.7} & \multirow{3}{*}{225.2} & \multirow{3}{*}{0.974} & \multirow{3}{*}{1.000}\\
     &  & Modify & 159.1 & 151.6 & 7.5 & 473.2 &  &  &  &  & \\
     &  & Reverse & 144.3 & 117.0 & 27.3 & 566.9 &  &  &  &  & \\
    \midrule
    \multirow{6}{*}{qwen3.8-27b} & \multirow{3}{*}{Off} & Forward & 185.9 & 184.0 & 1.8 & 168.0 & \multirow{3}{*}{153.6} & \multirow{3}{*}{13.2} & \multirow{3}{*}{166.7} & \multirow{3}{*}{0.455} & \multirow{3}{*}{0.200}\\
     &  & Modify & 123.8 & 118.9 & 4.8 & 398.7 &  &  &  &  & \\
     &  & Reverse & 119.0 & 94.3 & 24.6 & 519.8 &  &  &  &  & \\
    \cmidrule(l){2-12}
     & \multirow{3}{*}{On} & Forward & 74.5 & 67.3 & 7.1 & 442.0 & \multirow{3}{*}{159.0} & \multirow{3}{*}{4.6} & \multirow{3}{*}{163.6} & \multirow{3}{*}{1.000} & \multirow{3}{*}{1.000}\\
     &  & Modify & 90.0 & 83.9 & 6.1 & 483.0 &  &  &  &  & \\
     &  & Reverse & 103.2 & 81.3 & 21.8 & 536.8 &  &  &  &  & \\
    \midrule
    \multirow{6}{*}{deepseek-v4.1-flash} & \multirow{3}{*}{Off} & Forward & 64.5 & 62.0 & 2.4 & 178.1 & \multirow{3}{*}{181.3} & \multirow{3}{*}{13.5} & \multirow{3}{*}{194.8} & \multirow{3}{*}{0.480} & \multirow{3}{*}{0.206}\\
     &  & Modify & 28.3 & 21.4 & 6.9 & 464.9 &  &  &  &  & \\
     &  & Reverse & 56.1 & 28.4 & 27.7 & 520.4 &  &  &  &  & \\
    \cmidrule(l){2-12}
     & \multirow{3}{*}{On} & Forward & 21.6 & 14.1 & 7.3 & 451.1 & \multirow{3}{*}{202.5} & \multirow{3}{*}{5.1} & \multirow{3}{*}{207.6} & \multirow{3}{*}{1.000} & \multirow{3}{*}{1.000}\\
     &  & Modify & 28.3 & 21.4 & 6.9 & 481.8 &  &  &  &  & \\
     &  & Reverse & 47.0 & 20.0 & 26.9 & 540.8 &  &  &  &  & \\
    \midrule
    \multirow{6}{*}{gpt-6.1-sol} & \multirow{3}{*}{Off} & Forward & 47.5 & 43.2 & 4.3 & 398.4 & \multirow{3}{*}{102.4} & \multirow{3}{*}{2.7} & \multirow{3}{*}{105.2} & \multirow{3}{*}{1.000} & \multirow{3}{*}{1.000}\\
     &  & Modify & 23.8 & 16.8 & 7.0 & 570.4 &  &  &  &  & \\
     &  & Reverse & 50.9 & 25.3 & 25.5 & 576.2 &  &  &  &  & \\
    \cmidrule(l){2-12}
     & \multirow{3}{*}{On} & Forward & 37.6 & 31.0 & 6.5 & 443.3 & \multirow{3}{*}{105.3} & \multirow{3}{*}{2.2} & \multirow{3}{*}{107.5} & \multirow{3}{*}{1.000} & \multirow{3}{*}{1.000}\\
     &  & Modify & 25.6 & 19.0 & 6.6 & 571.6 &  &  &  &  & \\
     &  & Reverse & 47.5 & 24.2 & 23.3 & 582.1 &  &  &  &  & \\
    \midrule
    \multirow{6}{*}{claude-opus-5-5} & \multirow{3}{*}{Off} & Forward & 49.4 & 45.0 & 4.3 & 399.4 & \multirow{3}{*}{170.8} & \multirow{3}{*}{6.4} & \multirow{3}{*}{177.2} & \multirow{3}{*}{1.000} & \multirow{3}{*}{1.000}\\
     &  & Modify & 29.2 & 23.3 & 5.9 & 496.0 &  &  &  &  & \\
     &  & Reverse & 49.9 & 23.6 & 26.2 & 520.1 &  &  &  &  & \\
    \cmidrule(l){2-12}
     & \multirow{3}{*}{On} & Forward & 34.0 & 27.3 & 6.6 & 440.2 & \multirow{3}{*}{180.6} & \multirow{3}{*}{4.5} & \multirow{3}{*}{185.1} & \multirow{3}{*}{1.000} & \multirow{3}{*}{1.000}\\
     &  & Modify & 32.7 & 26.8 & 5.9 & 494.4 &  &  &  &  & \\
     &  & Reverse & 60.9 & 36.2 & 24.6 & 559.4 &  &  &  &  & \\
    \bottomrule
  \end{tabular*}
\end{table*}

\section{CFD Performance Measures}
\label{app:cfd-measures}
For the hydraulic cases in Table~\ref{tab:cfd-summary}, let $\bar p_{t,1}$ and $\bar p_{t,2}$ denote inlet and outlet total pressures averaged with the local volume flux as weight. The reported head and hydraulic efficiency are
\begin{equation}
 H_c=\frac{\bar p_{t,2}-\bar p_{t,1}}{\rho g},\qquad
 \eta_h=\frac{\rho g Q H_c}{P_{\rm rotor}} ,
\end{equation}
where $Q$ is in $\mathrm{m^3/s}$ and $P_{\rm rotor}$ is the power transferred to the fluid in watts. The head error is $|H_c-H_d|$ in metres. The streamlines in Fig.~\ref{fig:cfd-pump} use relative velocity in the rotating passage and absolute velocity in the stationary casing. Fig.~\ref{fig:cfd-fan-compressor} colors the aerodynamic streamlines by velocity magnitude to show the direction in each rotor.

\section{Rotor Assemblies}
\label{app:generated-assemblies}
\begin{figure*}[t]
    \includegraphics[width=0.95\linewidth]{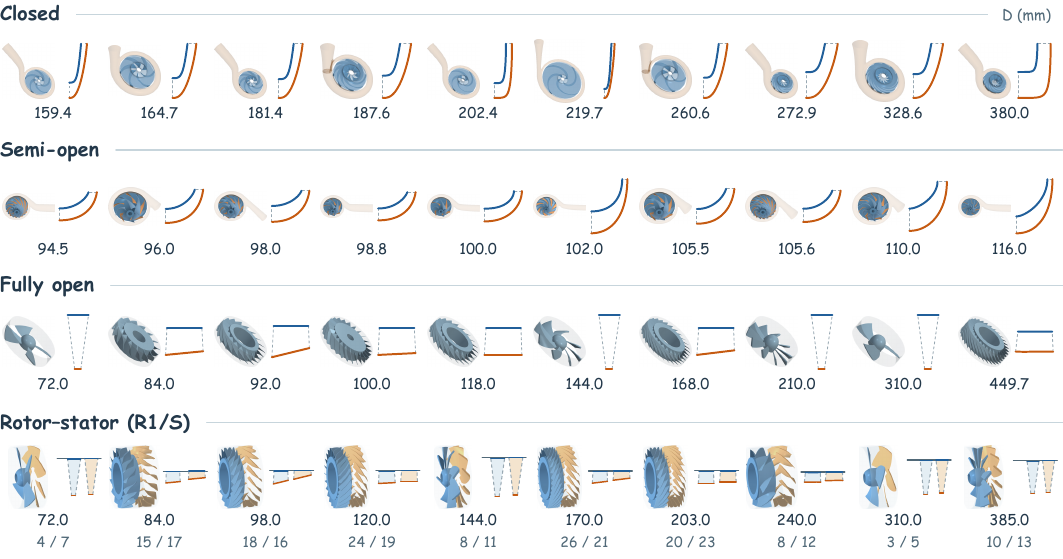}
    \caption{Rotor assemblies generated by VaneCraft and their corresponding meridional profiles.}
    \label{fig:generated-assemblies}
\end{figure*}

In Fig.~\ref{fig:generated-assemblies}, each assembly appears to the left of its meridional profile. The volutes have straight tangential or outward-curving radial outlets of different lengths and exit areas. Each row is ordered by the outer diameter $D$ of the rotor blades, measured about the shaft axis. In the bottom row, a single stator is fixed to the casing downstream of the fully open rotor. Its blades wrap around the axis in the opposite direction from the rotor blades.

\end{document}